\pdfoutput=1

\documentclass[conference]{IEEEtran}
\usepackage{cite}
\usepackage{amssymb,amsfonts}
\usepackage{algorithmic}
\usepackage{textcomp}
\usepackage{subcaption}

\ifCLASSINFOpdf
\usepackage{amsmath}
\usepackage{graphicx}
\graphicspath{ {Images/} }
  
\else

\fi

\begin{document}

\title{DroNet: Efficient Convolutional Neural Network Detector for Real-Time UAV Applications
\thanks{The dissemination of this work is being supported by the European Union's Horizon 2020 research and innovation programme under grant agreement No 739551 (KIOS CoE).}
}

\author{\IEEEauthorblockN{Christos Kyrkou, George Plastiras, \\ and Theocharis Theocharides}
\IEEEauthorblockA{KIOS Research and Innovation Center of Excellence\\
Department of Electrical and Computer Engineering\\
University of Cyprus \\
Nicosia, Cyprus\\
\{kyrkou.christos,gplast01,ttheocharides\}@ucy.ac.cy}
\and
\IEEEauthorblockN{Stylianos I. Venieris and Christos-Savvas Bouganis}
\IEEEauthorblockA{Department of Electrical and Electronic Engineering\\Imperial College London\\
\{stylianos.venieris10,christos-savvas.bouganis\}@imperial.ac.uk}
}

\maketitle

\begin{abstract}
Unmanned Aerial Vehicles (drones) are emerging as a promising  technology  for  both environmental and infrastructure monitoring, with broad use in a plethora of applications. Many such applications require the use of computer vision algorithms in order to analyse the information captured from an on-board camera. Such applications include detecting vehicles for emergency response and traffic monitoring. This paper therefore, explores the trade-offs involved in the development of a single-shot object detector based on deep convolutional neural networks (CNNs) that can enable UAVs to perform vehicle detection under a resource constrained environment such as in a UAV. The paper presents a holistic approach for designing such systems; the data collection and training stages, the CNN architecture, and the optimizations necessary to efficiently map such a CNN on a lightweight embedded processing platform suitable for deployment on UAVs. Through the analysis we propose a CNN architecture that is capable of detecting vehicles from aerial UAV images and can operate between 5-18 frames-per-second for a variety of platforms with an overall accuracy of $\sim95\%$. Overall, the proposed architecture is suitable for UAV applications, utilizing  low-power embedded processors that can be deployed on commercial UAVs. 
\end{abstract}

\IEEEpeerreviewmaketitle

\section{Introduction}
Deep learning (DL) has gathered significant interest recently as an Artificial Intelligence (AI) paradigm, with success in a wide range of applications such as image and speech recognition, autonomous systems, self-driving cars, cyber-physical systems, and many more. Among the most promising systems that can utilize deep learning are Unmanned Aerial Vehicles (UAVs) which are becoming an attractive solution for a wide range of applications. In particular, Road Traffic Monitoring (RTM), and Emergency Response (ER) systems constitute a domain where the use of UAVs is receiving significant interest. Under the above deployments, UAVs are responsible for searching, collecting and sending, in real time, vehicle information either for traffic regulation purposes or to aid search and rescue in emergency response.

In traffic monitoring applications, UAVs can perform vehicle identification, without the need for embedded sensors within cars and can be deployed in an area of interest at no additional cost; while for emergency response applications they greatly enhance remote sensing and situational awareness capabilities. 

A key challenge in the deployment of the above capabilities is the computing platform of a UAV is required to (1) consume minimal power, in order to minimize its effect on the battery and flight time of the system and (2) process the input data from its sensors with a low latency in order to make critical decisions, such as object avoidance and navigation, in real-time. Conventional on-board computing infrastructure consists mainly of general purpose machines, such as multi-core CPUs and low-power microcontrollers. The high computational workload of novel computer vision algorithms, such as convolutional neural networks, has led to the introduction of massively parallel architectures, with the prominence of Graphics Processing Units (GPUs), as accelerators \cite{Cavigelli2015}. To reach high performance, GPUs require the processing of inputs in batches in order to amortize the communication cost between the GPU and its external memory and achieve high throughput. Despite achieving high throughput, batch processing results in the deterioration of latency which makes GPUs inappropriate, in most cases, for the latency-sensitive tasks of a UAV. Moreover, the high power consumption of modern GPUs provide a high  overhead that can be prohibitive for low power UAVs.

An alternative to on-board processing are cloud-centric setups. In this scenario, the UAV collects data via its sensors and transmits them to a base server for analysis. Despite the fact that this case enables the UAV to save energy by off-loading compute intensive operations, the wireless transmission of a video feed can add significantly to the latency of the system. In latency- and security-sensitive tasks, the high latency and security risk of cloud computing may not be tolerated and thus local, on-board processing is necessary. Furthermore, in remote areas where there is no internet connection the detection process cannot be offloaded.

The high computational cost of video processing poses a challenge in mapping modern deep learning-based algorithms on low-cost, low-power computing platforms. Hence, in this paper we are concerned with developing algorithms based on deep learning for detecting vehicles, to make them efficient and suitable for real-time UAV applications running on embedded hardware platforms. We explore the parameter space in the design of convolutional neural networks as well as previously proposed models to identify an efficient architecture that can perform vehicle detection on images faster than previous works ($40\times$ speed-up) and at high accuracy ($\sim95\%$). The final proposed model is referred to as \textit{DroNet} and can be used as a starting point for developing UAV-based object detectors for various applications. 

\section{Background and Related Work}
\label{sec:background}

\subsection{Related Work on UAV-based Vehicle Detection}
\label{sec:UAVdet}
Object detection aims to find instances of objects from known classes in an image. This task typically yields the location and scale of the object in terms of a bounding box together with a probability on its class. UAV object detection has been extensively studied in the literature and traditional techniques utilize background subtraction \cite{de2015drone} to perform traffic estimation from static UAVs, or use Haar Cascade classifiers to detect vehicles \cite{AZEVEDO2014849}. The latest state-of-the-art techniques rely on deep convolutional neural networks (CNNs) \cite{NIPS2013_CNN_Obj_Dete}. For example a CNN is utilized in \cite{Chen2013VehDetSate} as a classifier to detect vehicles in grayscale images. However, such approaches need to process thousands of search windows and are thus inefficient for UAV platforms with limited hardware capabilities that need to operate near real-time. A more recent example is shown in \cite{audebert2017RS_CNNSegDet} where the authors utilize a deep learning framework that performs scene analysis of aerial images to first segment the image into various regions, and then extract the regions that correspond to vehicles and classify them into subcategories. The proposed network involves stacked deep neural networks for encoding and decoding the input image into segments and runs off-line on an NVIDIA Tesla GPU, which makes it unsuitable for a lightweight and low power UAVs. 

Contrary to existing work, in this work we target on-board processing on a UAV platform which may not be equipped with such high-end hardware. To this end, we propose a lightweight CNN architecture capable of running efficiently on embedded processors. Recent advances in object detection frameworks utilize deep learning and cast detection as a regression problem where the goal is to predict the location of bounding boxes in the image. Nevertheless, such techniques have not yet been exploited in UAV applications. In this work, we employ such techniques and tailor a single-shot CNN, that is trained on image data specifically for top-view vehicle detection, and is optimized to run on an embedded platform on-board a UAV. The next section outlines the basic principles of convolutional and single-shot detectors. 

\subsection{Convolutional Neural Network Detectors}
\label{sec:CNNs}
Convolutional neural networks (CNNs) are biologically inspired hierarchical models that can be trained to perform a variety of detection, recognition and segmentation tasks \cite{audebert2017RS_CNNSegDet}. CNNs are neural network architectures composed of multiple layers, with higher layers built on top of lower ones and capturing more abstract representations of the input data. The structure of a CNN typically comprises a feature extractor stage followed by a classifier. In the last decade, a lot of progress has been made on CNN-based object detection. Numerous object detectors have been proposed by the deep learning community, including Faster R-CNN \cite{FRCNN2017}, R-FCN \cite{RFCN2016}, YOLO \cite{Redmon2016} and SSD \cite{Liu2016}. The main emphasis of these designs is placed on improving (1) the detection accuracy and (2) the computational complexity of their methods in order to achieve real-time performance for mobile and embedded platforms \cite{LCDet2017}. The CNN-based object detectors can be taxonomized into two categories with respect to their high-level structure: 1) region-based detectors, usually consisting of a region-proposal stage followed by a classifier, and 2) single-shot detectors, which employ a single CNN to perform end-to-end object detection.

\subsubsection{Region-based Detectors}
Region-based detectors separate the prediction of the bounding box position from the object class prediction. A prominent example of such a classifier is Faster R-CNN \cite{FRCNN2017}, which divides its processing pipeline into two stages. The first stage, called the Region Proposal Network (RPN), employs the feature extractor of a CNN (e.g. VGG-16, ResNet, etc.) to process images and utilizes the output feature maps of a selected intermediate layer in order to predict bounding boxes of class-agnostic objects on an image. In the second stage, the box proposals are used to crop features of the same intermediate feature maps and pass them through a classifier in order to both predict a class and refine a class-specific box for each proposal. With typically hundreds proposals per image passed separately through the classifier, Faster R-CNN remains computationally heavy and poses a challenge in achieving high performance in embedded platforms.

\subsubsection{Single-Shot Detectors}
This class of detectors aims to avoid the performance bottlenecks of the 2-step region-based systems. The YOLO \cite{Redmon2016} framework casts object detection to a regression problem and in contrast to the RPN + classifier design of Faster R-CNN, employs a single CNN for the whole task. YOLO divides the input image into a grid of cells and for each cell outputs predictions for the coordinates of a number of bounding boxes, the confidence level for each box and a probability for each class. Compared to Faster R-CNN, YOLO is designed for real-time execution and by design provides a trade-off that favours high performance over detection accuracy. In addition, the open-source released version of YOLO has been developed using the C- and CUDA-based Darknet \cite{darknet13} framework, which enables the use of both CPUs and GPUs and is portable across a variety of platforms.

SSD \cite{Liu2016} is a single-shot detector aims to combine the performance of YOLO with the accuracy of region-based detectors. SSD extends the CNN architecture of YOLO by adding more convolutional layers and allowing the grid of cells for predicting bounding boxes to have a wider range of aspect ratios in order to increase the detection accuracy for objects of multiple scales. Despite the high detection accuracy and performance, the open-source released version of SSD has been developed using the Caffe framework. 

In this work, we focus on single-shot detectors due to their high performance and applicability to mobile and embedded systems. To this end, we select YOLO as our basis detection framework because the portability of its C-based release enables us to explore detection across diverse computing platforms.

\begin{figure*}[t]
  \includegraphics[width=\textwidth,height=4cm]{Models_cropped.pdf}
  \caption{Baseline Network Structures}
  \label{fig:structures}
\end{figure*}

\begin{figure*}[t]
	\centering
	\includegraphics[width=1\linewidth]{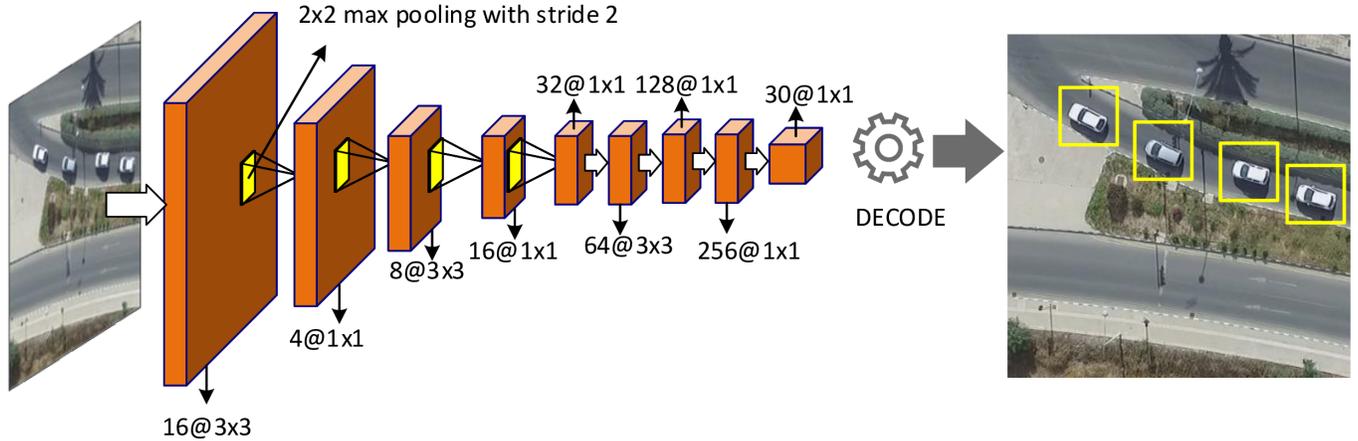}
	\caption{Architecture of the DroNet single shot CNN detector for top-view vehicle detection. It is comprised of  $3\times3$ and $1\times1$ convolutional layers and max-pooling layers that reduce the feature maps size by $2\times$}
	\label{fig:DroNetArch}
\end{figure*}

\section{Proposed Approach}
\subsection{Data Collection}
The training of the UAV-based single shot detector involves the collection of appropriate image data. To compile the training set, images were collected using a variety of methods including cropping of satellite images, retrieving images from the world-wide-web, and collecting urban traffic video footage from a UAV. By employing a variety of acquisition methods we were able to construct a dataset that captures vehicles under different conditions with regards to illumination, viewpoint, occlusion, color and type. In this way, after training, the detector would learn to identify vehicles in a variety of scenarios. Overall, a total of $350$ images where collected with a total of $\sim5000$ vehicles captured. 
\subsection{Training Process}
The produced image database was used to train the networks under consideration. The training process involved annotating all vehicles in the images. We annotated only vehicles with $50\%$ of their body visible within the image. A number of CNN models were designed and trained using the Darknet framework on an NVIDIA Titan Xp GPU. The different models varied in terms of number of layers, filter sizes and input image size. All models were trained using the loss function defined in \cite{Redmon2016} to jointly predict whether vehicles were detected together with the bounding boxes of vehicles in the image.

\subsection{CNN Models}
While there has been extensive investigation on reducing the complexity of well studied CNN models in the form of parameter compression and quantization \cite{Rastegari2016}, there has been limited effort on developing specialized CNN designs for UAVs
which need radical changes due to highly resource-constrained environment both in computer power and memory bandwidth. As such, in this work we focus on designing an efficient and lightweight network to accelerate the execution of the model with minimal compromise on the achieved accuracy.
The \textit{Tiny-YOLO}, model is a smaller model of YOLO detection that is fast and works efficiently on a desktop-class GPU \cite{Redmon2016}. First, we adapted this basic model to detect only one class, in our case top-view vehicles. Second, we explore the impact on performance by changing the structure of a CNN network such as the number of filters, the number of layers, the image size, the number of convolution and the pooling layers. Overall we design four different structures (\textit{SmallYoloV3}, \textit{TinyYoloVoc}, \textit{TinyYoloNet}, and \textit{DroNet}) all shown in \text{Fig. \ref{fig:structures}}, with different parameters including the layers and the type of each layer (conv,maxpool,detection) together with the configuration of the layers in terms of the number of filters, the size of filters in each layer and the input and the output size of the feature maps. The overall design approach is as follows:

\subsubsection{Number and Size of Filters}
We use the structure of \textit{Tiny-YOLO} model \cite{Redmon2016} as starting point and from there we decrease the number of filters in each layer in order to get a smaller network which can lead to a faster detection. To reduce the amount of operations per input, we reduced the number of filters by decreasing the number of filters per layer and coarsely pruning whole layers. For the convolutional layers we gradually increase the number of filters up to a point as the net gets deeper so that we keep the compute requirements low. In total there are $9$ convolutional layers in the models shown in Fig. \ref{fig:structures}, with the max-pooling layers ranging between $4-6$. 

\subsubsection{Input Image Size}
The size of the input image that is processed by the network is a factor that can affect both accuracy and performance. On the one hand, larger images often lead to higher detection accuracy by the inclusion of more information. Nevertheless, larger images also lead to larger feature maps that have to be processed in the network and hence greater computational and memory load. In our experiments, we varied the image size in both the training and testing stages from $352$ to $608$. 

\subsection{Application Level Optimizations}
When the UAV platform is capable of providing altitude information we can incorporate this into the detection process by restricting the possible sizes of detected objects. For example, vehicles can appear within a certain range based on the UAV altitude; any objects that are not within this range can be discarded as false detections, based on their size and feasibility with respect to the UAV altitude and real object size. However, this is left as a future work as it is complementary to this work.

\section{Evaluation and Experimental Results}
In this section, we present a comprehensive quantitative evaluation of the four CNN structures. The basic network models are trained and tested for various input size using the constructed vehicle dataset. Throughout the experiments, three platforms were targeted: (1) an Intel CPU i5-2520m at 3.20 GHz having two cores with two hardware threads per core and 3 MB cache, (2) an Odroid UX4 with an octacore Samsung Exynos-5422 CPU at 2GHz with 2 GB of RAM and (3) Rasperry Pi 3 containing a quad-core ARM Cortex-A53 at 1.20 GHz with 1 GB of RAM. The reason for choosing different platforms is that UAV platforms can differ in there implementation depending on the use-case and deployment stratgy; from very lightweight with only on-board components to requiring dedicated computing infrastructure at the ground station.

\noindent
\textbf{Metrics:} To evaluate the performance of each model, we employ the following four metrics:
\subsubsection{Intersection over Union (IoU)}
In the context of object detection, the IoU metric captures the similarity between the predicted region and the ground-truth region for an object present in the image. This metric is defined as the size of the intersection of the predicted and ground-truth regions divided by their union.
\subsubsection{Sensitivity}
This metric is defined as the proportion of true positives that are correctly identified by the detector. This metric is calculated by taking into account the True Positives ($T_{pos}$) and False Negatives ($F_{neg}$) of the detected cars as given by (\ref{sensitivity_equ}).

\begin{equation}
    \label{sensitivity_equ}
    Sensitivity = \frac{T_{pos}}{T_{pos}+F_{neg}}
\end{equation}

\subsubsection{Precision}
This metric is a widely used metric by the object detection community and is defined as the proportion of True Positives among all the detections of the system as captured by (\ref{precision_equ}).

\begin{equation}
    \label{precision_equ}
    Precision = \frac{T_{pos}}{T_{pos} + F_{pos}}
\end{equation}

\subsubsection{Frames-Per-Second (FPS)}
The rate at which an object detector is capable of processing incoming camera frames.

\noindent
\textbf{Score:}
In order to capture the overall performance of each detector, we define a composite \textit{score} metric. This metric consists of a linear combination of IoU, Sensitivity and Precision together with the achieved FPS on a particular platform. We parametrize the score with respect to a vector of weights $\mathrm{w} \in [0,1]^4 $ as given by (\ref{score_equ}),

\begin{align}
    \label{score_equ}
    Score(\boldsymbol{\mathrm{w}}) &= \mathrm{w_1} \times FPS + \mathrm{w_2} \times IoU \\
    &+ \mathrm{w_3} \times Sensitivity + \mathrm{w_4} \times Precision \nonumber \\
    \text{subject to} \sum^4_{i=1} \mathrm{w_i} &= 1 \nonumber
\end{align}
where each weight captures the application-level importance of each metric. Considering that in our case we target real-time applications we assigned weights on these metrics to capture the best trade-off between a fast detector and good accuracy. We prioritized FPS with a weight of 0.4 over the other three accuracy-related metrics, which were equally weighted with 0.2. Hence, the model with the highest overall score will be the one that is best suited for our application demands and platform constraints.

\subsection{Parameter Space Exploration on CPU Platform}

\subsubsection{Number and Size of Filters}

In this section, we present the effect of modifying the architectural parameters of the underlying networks on detection accuracy and performance on a CPU platform. The networks vary with respect to their topology and the configuration of the layers. The presented results are normalized, between CNNs by first dividing them with the maximum value of each metric across all CNNs, in order to lie in the range $[0,1]$. Thus making it easier to compare and contrast the different models and their trade-offs.

\begin{figure}[h]
	\centering
	\begin{subfigure}{.7\linewidth}
		\includegraphics[width=1\columnwidth]{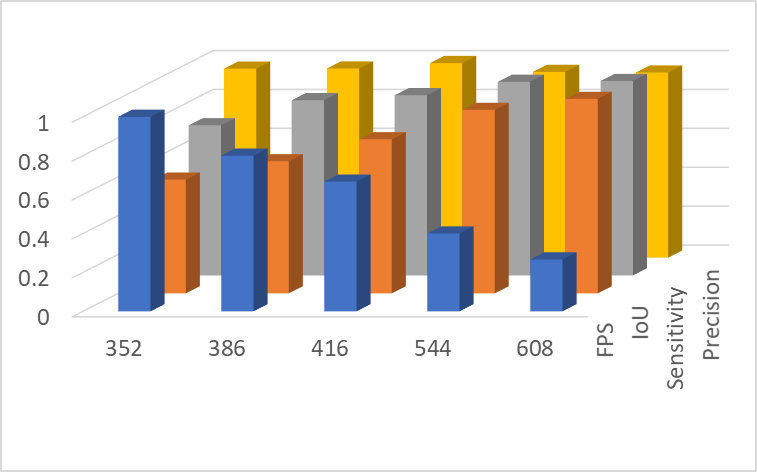}
		\label{fig:sim_t1}
	\end{subfigure}
	\hskip2em
	\begin{subfigure}{.7\linewidth}
		\includegraphics[width=1\columnwidth]{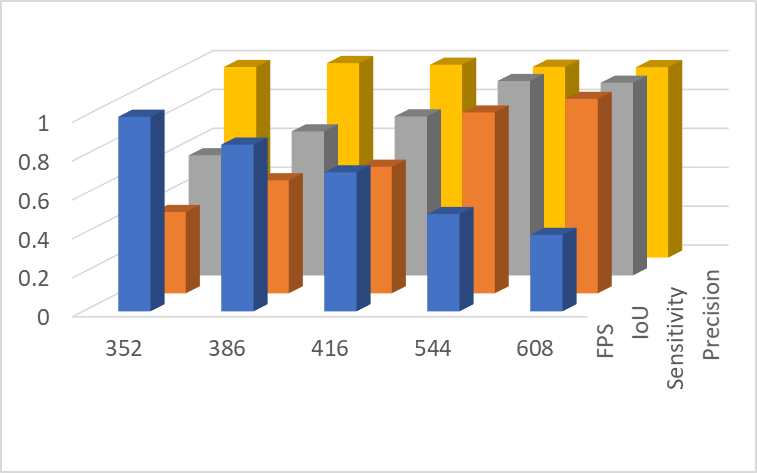}
		\label{fig:sim_t2}
	\end{subfigure}
	\begin{subfigure}{.7\linewidth}
		\includegraphics[width=1\columnwidth]{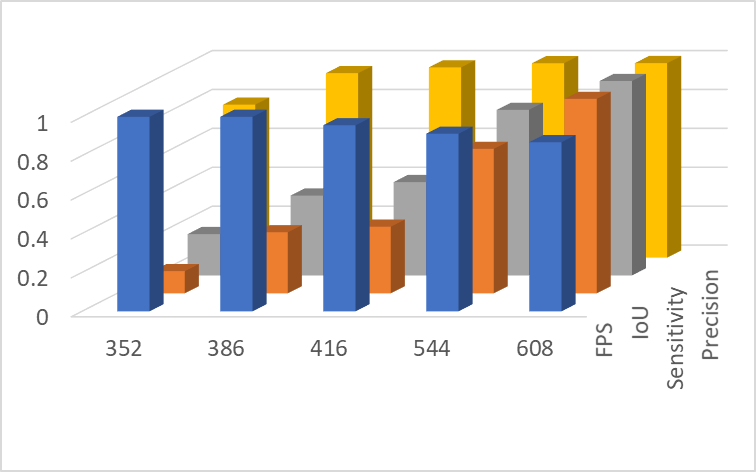}
		\label{fig:sim_t3}
	\end{subfigure}
	\begin{subfigure}{.7\linewidth}
		\includegraphics[width=1\columnwidth]{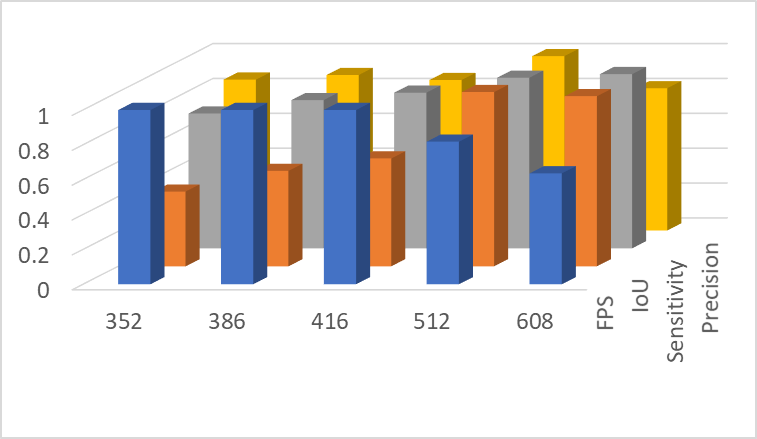}
		\label{fig:sim_t4}
	\end{subfigure}
	\caption{Normalized performance metrics for basic model configurations for different input image sizes: (a) TinyYoloVoC (b) SmallYoloV3 (c) SmallYoloV3 (d) DroNet.}
	\label{fig:simulation}
\end{figure}

Fig. {\ref{fig:simulation}} shows the performance comparison between the four models. In our test set, with $386\times386$ as image size, \textit{TinyYoloNet} achieved $10\times$ higher performance than \textit{TinyYoloVoc} with decreased detection sensitivity and precision by $20\%$ and $10\%$ respectively and a drop in IoU of $0.11$. The  network \textit{SmallYoloV3}, with $386\times386$ as image size achieved the highest frame-rate among all network designs with $23$ FPS. Nevertheless, the substantial reduction in the number of weights led to a decrease in sensitivity which was $53\%$ lower, which prohibits us from using it for robust vehicle detection.

\begin{figure}[t]
\centering
\includegraphics[width=0.8\columnwidth]{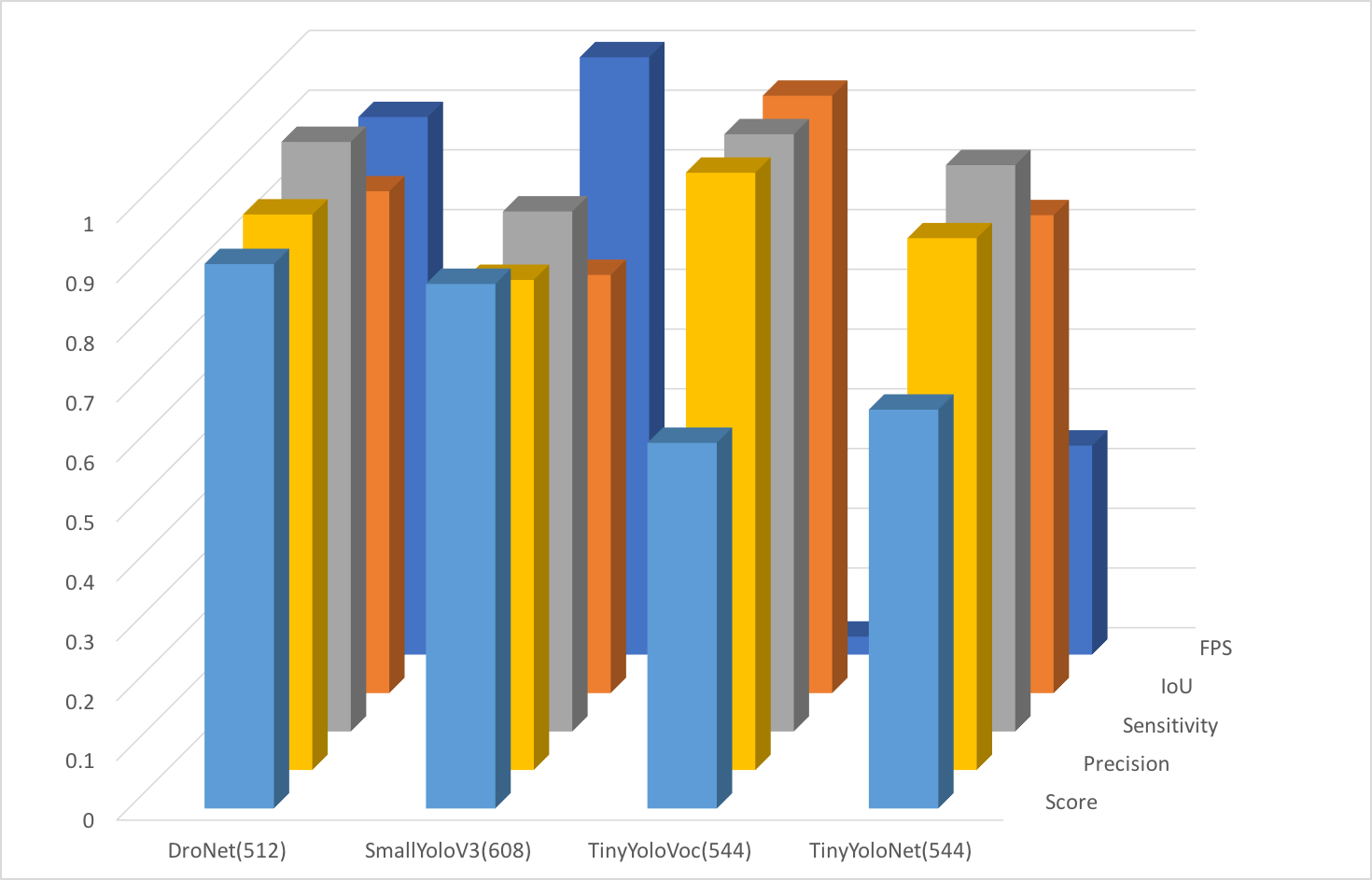}
\caption{Combined metric results for the best models using the weighted Score metric.}
\label{fig:OverallResult}
\end{figure}

Fig. \ref{fig:simulation} shows the significant performance gains starting from \textit{TinyYoloVoc} to \textit{TinyYoloNet} followed by \textit{SmallYoloV3} and then to \textit{DroNet}. For example, comparing these models for the same input size, $386$, the performance of \textit{DroNet} is $30\times$ faster compared to \textit{TinyYoloVoc} with a minimal drop of $0.08$ on the IoU. Moreover, the detection sensitivity and precision were also affected in a limited manner with a decrease of $2\%$ and $6\%$ respectively. With respect to the weighted score metric, \textit{DroNet} achieved a $3\%$ increase over \textit{TinyYoloVoc} due to its large speed up. Furthermore, we observe a slight increase in false detections of \textit{DroNet}. This is a symptom of the loss of information during training that was caused by the decrease in the size and number of layers. 

\subsubsection{Input Image Size}
By increasing the input image size, \textit{TinyYoloVoc} demonstrated an increase of $0.17$ in the IoU and $10\%$ in the sensitivity at the cost of $3\times$ slower performance. \textit{TinyYoloVoc} with increase size of input images was the model configuration with the highest accuracy on our test set, reaching $97\%$.
Fig. \ref{fig:simulation} presents the effect of different input image sizes in the range $352$ to $608$ on the metrics of our faster models. By using images of larger size, the detection sensitivity is increased by an average of $1.28\times$ across the models. Conversely, the larger input size deteriorates performance with an average of $0.81\times$ across the models. A trade-off between higher detection accuracy and acceptable performance can be reached by applying inputs with a size between $416$ and $544$. In this range, the accuracy gain is higher than the penalty on detection speed. 

Fig. \ref{fig:OverallResult} shows the best performance of each one of the models on our dataset. In this context, based on the explored design space, a size of $512\times512$ maximizes the weighted score metric of the $DroNet$ model and therefore we select it for use on the UAV platform.

\begin{figure}[t]
	\centering
	\begin{subfigure}{1\linewidth}
		\includegraphics[width=1\columnwidth]{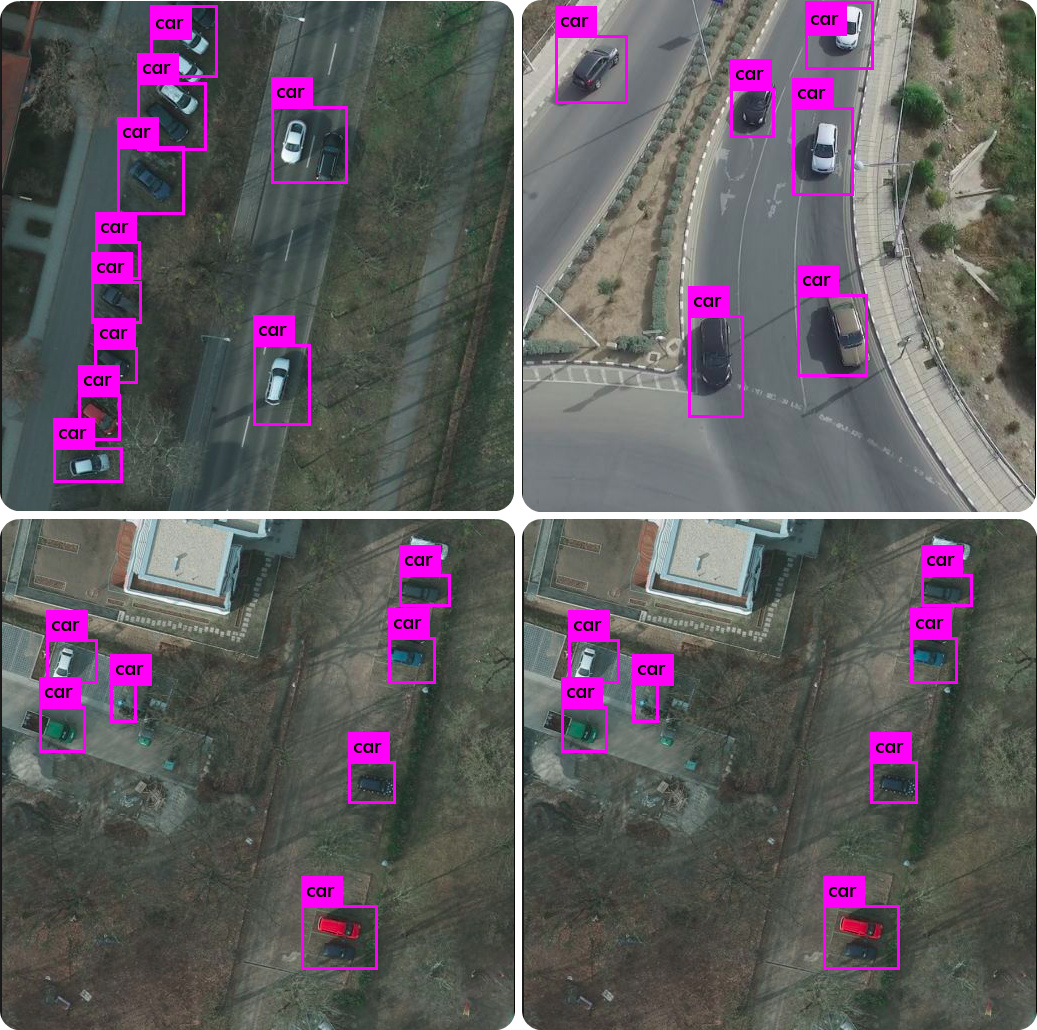}
		\label{fig:sim_t1}
	\end{subfigure}
	\hskip2em
	\begin{subfigure}{1\linewidth}
		\includegraphics[width=1\columnwidth]{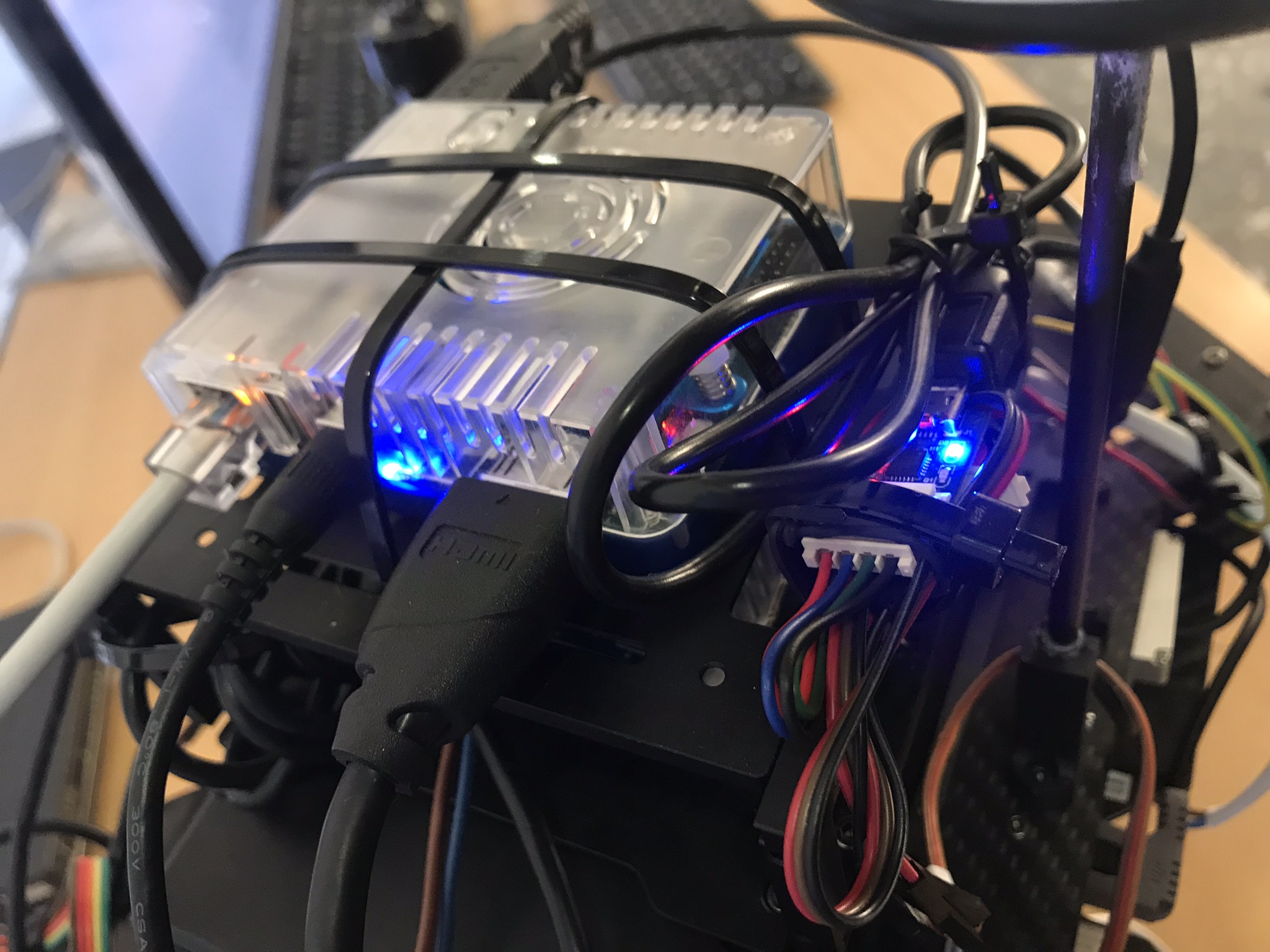}
		\label{fig:sim_t2}
	\end{subfigure}
	
	\caption{Evaluation on UAV Platform: (a) Detection results on images acquired from a UAV and from aerial databases. (b) Odroid board attached on DJI Matrice 100 UAV.}
	\label{fig:UAVplatform}
\end{figure}

\subsection{DroNet on UAV Platform}

In this section we implement the \textit{DroNet} model on a DJI Matrice 100 UAV by interfacing it with two different computing platforms. 

\subsubsection{Odroid-XU4}
To map our proposed model on Odroid-XU4 platform we used Darknet\cite{darknet13} to implement our network structure. We first install Darknet for Odroid-XU4 and loaded the pretrained \textit{DroNet} weights. We use the on board camera to retrieve real time video feed and pass it frame by frame to the processing board where the vehicles are detected (\ref{fig:UAVplatform}). Detection results of the \textit{DroNet} model from the UAV are shown in Fig. \ref{fig:UAVplatform}.
Odroid performance was around $8-10$ FPS with the accuracy maintained around $95\%$. The performance was decreased due to the fact that the detection was spread uniformly across all cores with $50\%$ utilization for each one. This was a problem that we were facing with Darknet implementation but it may be possible to optimize implementation and employ the use of all CPU cores with $100\%$ usage in the future, which may result in further improvement of performance. It is worth noticing that the proposed \textit{DroNet} model was $40\times$ faster than \textit{TinyYoloVoc} performance which achieved only $0.1$ FPS on Odroid.
\subsubsection{Raspberry Pi 3 - Model B}
Moreover, we also implemented our proposed model on Raspberry Pi 3. The accuracy was maintained again around $95\%$ but the performance was only $5-6$ FPS this time mainly due to the less capable chipset on this platform. However, for certain applications which exhibit slow varying dynamics this option can be a viable one.

\section{Conclusion \& Future Work}
This paper presented a design exploration for single-shot convolutional neural network detectors for UAV-based vehicle detection. Specifically, a dataset was constructed and different models were trained and evaluated for different configurations and metrics. The resulting CNN referred as \textit{DroNet} is capable of $5-18$ FPS on different platforms while achieving $95\%$ detection accuracy. Potential future work includes the performance improvements by applying finer-level optimizations to reduce bitwidth precisions.  Finally, we also aim to significantly enhance the training set with additional images and object classes (e.g., pedestrians, motorbikes, etc.) necessary for different UAV applications such as emergency response, traffic management, infrastructure inspection, . 

\section*{Acknowledgment}

Christos Kyrkou gratefully acknowledges the support of NVIDIA Corporation with the donation of the Titan Xp GPU used for this research.

\bibliographystyle{IEEEtran}

\end{document}